# Stable Independence and Complexity of Representation

**Peter de Waal** and **Linda C. van der Gaag**

Institute of Information and Computing Sciences, Utrecht University

P.O. Box 80089, 3508 TB Utrecht, the Netherlands

{waal,linda}@cs.uu.nl

## Abstract

The representation of independence relations generally builds upon the well-known semi-graphoid axioms of independence. Recently, a representation has been proposed that captures a set of dominant statements of an independence relation from which any other statement can be generated by means of the axioms; the cardinality of this set is taken to indicate the complexity of the relation. Building upon the idea of dominance, we introduce the concept of stability to provide for a more compact representation of independence. We give an associated algorithm for establishing such a representation. We show that, with our concept of stability, many independence relations are found to be of lower complexity than with existing representations.

## 1  Introduction

The concept of independence plays a key role in probabilistic systems, since effective use of knowledge about independences allows these systems to deal with the computational complexity of their problem-solving tasks. An independence relation on the set of variables of such a system is a complete description of the independences among the variables concerned. It thus captures all independences conditional on any possible available evidence and therefore specifies all independences that could possibly arise. In view of a specific problem-solving process, at any time during reasoning, only some of the independences from the relation apply. These are the independences that pertain to the current context of available evidence.

The concept of independence has been a subject of extensive studies. Pearl and his co-researchers were among the first to formalise properties of independence in an axiomatic system and to develop a logic for independence [4, 5]. Their semi-graphoid axioms provide for comput-ing new independence statements from a basic set of statements and allow for verifying whether a new statement logically follows from a given set of independence statements.

The representation of independence relations generally builds upon the semi-graphoid axioms of independence. The basic idea is to capture a number of statements from a relation explicitly and let the other statements be defined implicitly by the axioms. Recently, Studený [7] proposed a new representation based upon this idea, that captures the so-called dominant statements of a relation. He further introduced a concept of complexity for independence relations that is defined as the least cardinality of a generating set of statements.

In this paper we further elaborate on the idea of capturing an independence relation by its dominant statements. We introduce the concept of stability of independence and say that two sets of variables are stably independent if they are independent in the current context of available evidence and remain to be so as the context grows. We then show that by exploiting the concept of stability, a substantial reduction in size of the set of dominant statements for a given relation can be achieved.

The paper is organised as follows. In Section 2, we briefly review the semi-graphoid axioms of independence. In Section 3, we introduce our concept of stability. In Section 4, we address the representation of an independence relation by means of sets of dominant statements. We present an algorithm for establishing such a set based upon our concept of stability in Section 5. We address the complexity of independence relations in Section 6. The paper ends with our concluding observations in Section 7.

## 2  Independence Revisited

We consider a finite index set $N = \{1, \ldots, n\}$, $n \geq 1$, where each index denotes a statistical variable. The set of ordered triplets $\langle A, B | C \rangle$ of pairwise disjoint subsets of $N$, where $A$ and $B$ are non-empty, is denoted by $\mathcal{T}(N)$. The *symmetric image* of a triplet $u = \langle A, B | C \rangle$ is the triplet $\langle B, A|C \rangle$; it is denoted by $\mathrm{sym}(u)$. For simplicity of notation we will often write $AB$ to denote the union $A \cup B$. We will also use the notation $\mathcal{I}\langle A, B|C \rangle$ to indicate $\langle A, B|C \rangle \in \mathcal{I}$. A triplet $\langle A, B|C \rangle$ will be taken to denote that $A$ and $B$ are independent given $C$.

We review the four basic axioms of independence [4].

**Definition 2.1** *A ternary relation $\mathcal{I}$ on $N$ is a* semi-graphoid independence relation, *or semi-graphoid for short, if it satisfies the following four axioms:*

**A1:** $\mathcal{I}\langle X, Y|Z \rangle \rightarrow \mathcal{I}\langle Y, X|Z \rangle$;

**A2:** $\mathcal{I}\langle X, YW|Z \rangle \rightarrow \mathcal{I}\langle X, Y|Z \rangle \wedge \mathcal{I}\langle X, W|Z \rangle$;

**A3:** $\mathcal{I}\langle X, YW|Z \rangle \rightarrow \mathcal{I}\langle X, Y|WZ \rangle$;

**A4:** $\mathcal{I}\langle X, Y|Z \rangle \wedge \mathcal{I}\langle X, W|YZ \rangle \rightarrow \mathcal{I}\langle X, YW|Z \rangle$;

*for all sets of variables $X$, $Y$, $Z$, $W \subset N$.*

For a semi-graphoid independence relation $\mathcal{I}$ these axioms with each other convey the idea that learning irrelevant information does not alter the independences among the variables discerned. The four axioms are termed the *symmetry* (A1), *decomposition* (A2), *weak union* (A3), and the *contraction axiom* (A4), respectively. The axioms have been proven logically independent [4]. The term semi-graphoid refers to the representation of independence relations in graphical structures [3, 4]. The axioms are therefore sometimes referred to as the *semi-graphoid axioms*.

**Definition 2.2** *Let $\mathcal{I}$ be a ternary relation on $N$. Then,* $\mathrm{sem}(\mathcal{I})$ *is the closure of $\mathcal{I}$ under the semi-graphoid axioms, that is,*

$$\mathrm{sem}(\mathcal{I}) = \bigcap_{\substack{\mathcal{I} \subset \mathcal{M} \subset \mathcal{T}(N) \\ \mathcal{M} \text{ is a semi-graphoid}}} \mathcal{M}.$$

For a given set $\mathcal{I}$ of triplets, $\mathrm{sem}(\mathcal{I})$ thus is the set of all triplets in $\mathcal{T}(N)$ that can be derived by application of the semi-graphoid axioms to the elements of $\mathcal{I}$. Note that for any $\mathcal{I} \subset \mathcal{T}(N)$, $\mathrm{sem}(\mathcal{I})$ is a semi-graphoid independence relation. Also $\mathrm{sem}(\mathcal{I}) = \mathcal{I}$ iff $\mathcal{I}$ is a semi-graphoid.

In [6] Studený defined a concept of dominance for triplets.

**Definition 2.3** *Let $\langle T, U|W \rangle$, $\langle X, Y|Z \rangle \in \mathcal{T}(N)$. We say that $\langle X, Y|Z \rangle$ dominates $\langle T, U|W \rangle$, denoted $\langle T, U|W \rangle \prec \langle X, Y|Z \rangle$, if $T \subset X$, $U \subset Y$, and $Z \subset W \subset XYZ$. Let $\mathcal{I}$ be a ternary relation on $N$. A triplet in $\mathcal{I}$ that is not dominated by any other triplet in $\mathcal{I}$ is termed* maximally dominant *in $\mathcal{I}$.*

We have from the definition that $u \prec v$ iff $u$ can be derived from $v$ by application of the symmetry, decomposition, and weak union axioms, or equivalently, if $u \in \mathrm{sem}(\{v\})$. This observation generalises to the following lemma.

**Lemma 2.4** *Let $\mathcal{I} \subset \mathcal{T}(N)$ be a semi-graphoid independence relation on $N$. Let $\mathcal{D} \subset \mathcal{I}$ be the set of all triplets that are maximally dominant in $\mathcal{I}$. Then,* $\mathrm{sem}(\mathcal{D}) = \mathcal{I}$.

*Proof.* Let $v \in \mathcal{I}$. Then there exists a $u \in \mathcal{D}$, such that $v \prec u$, and we thus have $v \in \mathrm{sem}(\{u\}) \subset \mathrm{sem}(\mathcal{D})$, from which we conclude $\mathcal{I} \subset \mathrm{sem}(\mathcal{D})$. Since $\mathcal{D} \subset \mathcal{I}$, we further have $\mathrm{sem}(\mathcal{D}) \subset \mathrm{sem}(\mathcal{I}) = \mathcal{I}$. $\square$

# 3 Stable and Unstable Independence

An independence relation can be viewed as a *static* description of the independences among the variables concerned, unrelated to any specific reasoning process. At any time during reasoning, however, only some of the triplets from the relation apply to the current situation that is described by the set of variables for which information is available. As inference progresses, learning new information causes the set of relevant triplets to change dynamically. Some of the independences, however, will remain to hold. We say that these independences are stable.

**Definition 3.1** *Let $\mathcal{I} \subset \mathcal{T}(N)$ be a semi-graphoid independence relation on $N$. Then,*

- *a triplet $\mathcal{I}\langle X, Y|Z \rangle$ is called* stable *in $\mathcal{I}$ if $\mathcal{I}\langle X, Y|Z' \rangle$ for all sets $Z'$ with $Z \subset Z'$; if $XYZ = N$, then $\mathcal{I}\langle X, Y|Z \rangle$ is called* trivially stable;

- *a triplet $\mathcal{I}\langle X, Y|Z \rangle$ is called* unstable *in $\mathcal{I}$ if it is not stable in $\mathcal{I}$.*

*The set of all triplets that are stable in $\mathcal{I}$ is called the* stable part *of $\mathcal{I}$, and will be denoted by $\mathcal{S}_\mathcal{I}$; the set of all unstable triplets is called the* unstable part *of $\mathcal{I}$, denoted $\mathcal{U}_\mathcal{I}$.*

We use the notation $\mathcal{S}_\mathcal{I}\langle A, B|C \rangle$ to denote $\langle A, B|C \rangle \in \mathcal{S}_\mathcal{I}$.

The stable part of an independence relation has a highly regular structure. In this section, we state various properties of stable independence that we will exploit in the sequel. We begin by observing that the stable part of a semi-graphoid independence relation adheres to the four semi-graphoid axioms.

**Lemma 3.2** *Let $\mathcal{I} \subset \mathcal{T}(N)$ be a semi-graphoid independence relation on $N$ and let $\mathcal{S}_\mathcal{I}$ be its stable part. Then, $\mathcal{S}_\mathcal{I}$ satisfies the axioms*

**S1:** $\mathcal{S}_\mathcal{I}\langle X, Y|Z \rangle \rightarrow \mathcal{S}_\mathcal{I}\langle Y, X|Z \rangle$;

**S2:** $\mathcal{S}_\mathcal{I}\langle X, YW|Z \rangle \rightarrow \mathcal{S}_\mathcal{I}\langle X, Y|Z \rangle \wedge \mathcal{S}_\mathcal{I}\langle X, W|Z \rangle$;

**S3:** $\mathcal{S}_\mathcal{I}\langle X, YW|Z \rangle \rightarrow \mathcal{S}_\mathcal{I}\langle X, Y|WZ \rangle$;

**S4:** $\mathcal{S}_\mathcal{I}\langle X, Y|Z \rangle \wedge \mathcal{S}_\mathcal{I}\langle X, W|YZ \rangle \rightarrow \mathcal{S}_\mathcal{I}\langle X, YW|Z \rangle$;

*for all sets of variables $X$, $Y$, $Z$, $W \subset N$.*

The proof of the lemma is a straightforward application of Definition 3.1 and is therefore omitted.

From the previous lemma, we have that the stable part of a semi-graphoid independence relation is a semi-graphoid independence relation by itself. As a consequence, there exist semi-graphoid independence relations of which the unstable part is empty; such relations are termed *ascending* [2].

In addition to the properties reviewed so far, the stable part of an independence relation satisfies the property of *strong union* stated in the following lemma [4]. The lemma follows directly from the definition of stable independence.

**Lemma 3.3** *Let $\mathcal{I} \in \mathcal{T}(N)$ be a semi-graphoid independence relation on $N$ and let $\mathcal{S}_{\mathcal{I}}$ be its stable part. Then, $\mathcal{S}_{\mathcal{I}}$ satisfies the axiom*

**S5:** $\mathcal{S}_{\mathcal{I}}\langle X, Y | Z \rangle \rightarrow \mathcal{S}_{\mathcal{I}}\langle X, Y | ZW \rangle$

*for all sets of variables $X$, $Y$, $Z$, $W \subset N$.*

We note that the weak union axiom is implied for stable independence by the strong union and the decomposition axioms. The strong union axiom for stable independence now implies the following property.

**Lemma 3.4** *Let $\mathcal{I} \in \mathcal{T}(N)$ be a semi-graphoid independence relation on $N$ and let $\mathcal{S}_{\mathcal{I}}$ be its stable part. Then,*

$$\mathcal{S}_{\mathcal{I}}\langle X, Y | Z \rangle \wedge \mathcal{S}_{\mathcal{I}}\langle X, W | Z \rangle \rightarrow \mathcal{S}_{\mathcal{I}}\langle X, YW | Z \rangle$$

*for all sets of variables $X, Y, Z, W \subset N$.*

*Proof.* We assume that $\mathcal{S}_{\mathcal{I}}\langle X, Y | Z \rangle$ and $\mathcal{S}_{\mathcal{I}}\langle X, W | Z \rangle$ for some sets of variables $X, Y, Z, W \subset N$. From $\mathcal{S}_{\mathcal{I}}\langle X, W | Z \rangle$, we find that $\mathcal{S}_{\mathcal{I}}\langle X, W | YZ \rangle$ by the strong union property for stable independence. From $\mathcal{S}_{\mathcal{I}}\langle X, Y | Z \rangle$ and $\mathcal{S}_{\mathcal{I}}\langle X, W | YZ \rangle$ we conclude $\mathcal{S}_{\mathcal{I}}\langle X, YW | Z \rangle$ by the contraction property. □

From the property stated in the previous lemma, we have that the decomposition axiom actually is a bi-implication for stable independence. The property from Lemma 3.4 is therefore sometimes referred to as the *composition* property for stable independence [4]. In the sequel we refer to the five axioms S1–S5 as the *stable semi-graphoid axioms*.

Analogous to the definitions of semi-graphoid closure and dominance we now define the stable semi-graphoid closure and the concept of stable dominance.

**Definition 3.5** *Let $\mathcal{I}$ be a ternary relation on $N$. Then, $\mathrm{stab}(\mathcal{I})$ is the closure of $\mathcal{I}$ under the stable semi-graphoid axioms, that is,*

$$\mathrm{stab}(\mathcal{I}) = \bigcap_{\substack{\mathcal{I} \subset \mathcal{M} \subset \mathcal{T}(N) \\ \mathcal{M} \text{ is a stable semi-graphoid}}} \mathcal{M}.$$

From the definition we have that $\mathrm{stab}(\mathcal{I}) = \mathcal{I}$ iff $\mathcal{I}$ is a stable semi-graphoid independence relation.

**Definition 3.6** *Let $\langle T, U | W \rangle$, $\langle X, Y | Z \rangle \in \mathcal{T}(N)$. We say that $\langle X, Y | Z \rangle$ s-dominates $\langle T, U | W \rangle$, denoted $\langle T, U | W \rangle \prec\!\!\prec \langle X, Y | Z \rangle$, if $T \subset X$, $U \subset Y$, and $Z \subset W$. Let $\mathcal{I}$ be a ternary relation on $N$. A triplet that is not s-dominated by any other triplet in $\mathcal{I}$ is termed* maximally s-dominant *in $\mathcal{I}$.*

We have from the definition that $u \prec\!\!\prec v$ iff $u$ can be derived from $v$ by application of the symmetry, decomposition and strong union axioms. Note that we define the concept of s-dominance for arbitrary triplets; it is not restricted to stable semi-graphoids. In the remainder of this paper we shall refer to ordinary dominance, as defined in Definition 2.3, by the term *o-dominance*. From their definitions it is immediate that o-dominance implies s-dominance.

**Lemma 3.7** *For all $u, v \in \mathcal{T}(N)$, if $u \prec v$, then $u \prec\!\!\prec v$.*

The reverse property does not hold. Consider for instance $\langle X, W | YZ \rangle \prec\!\!\prec \langle X, W | Y \rangle$. The concept of s-dominance now is related to the stable semi-graphoid closure in the same way as o-dominance is related to the semi-graphoid closure.

**Lemma 3.8** *Let $\mathcal{I} \subset \mathcal{T}(N)$ be a stable semi-graphoid independence relation on $N$. Let $\mathcal{D} \subset \mathcal{I}$ be the set of all triplets that are maximally s-dominant in $\mathcal{I}$. Then, $\mathrm{stab}(\mathcal{D}) = \mathcal{I}$.*

The proof is analogous to that of Lemma 2.4.

## 4    Representation of Independence

In his work on complexity of representation Studený [7] exploits the concept of o-dominance to construct a representation of an independence relation that is more efficient than a list of the relation's triplets. His representation is based on the idea of Lemma 2.4 that an independence relation $\mathcal{I}$ is uniquely determined by its set $\mathcal{D}$ of o-dominant triplets. Typically $\mathcal{D}$ contains fewer triplets than $\mathcal{I}$. In this paper we further elaborate on this idea of representing an independence relation by its dominant triplets and allow s-dominant triplets to represent part of $\mathcal{I}$. We shall show that this leads to an even smaller number of triplets to represent $\mathcal{I}$.

In order to construct dominant triplets from a given set of triplets we introduce two operators. The $\star$-operator for constructing a (potentially) new o-dominant triplet for the semi-graphoid closure of two independence statements $u$ and $v$ was defined by Studený.

**Definition 4.1** *Let $u = \langle A, B | C \rangle$, $v = \langle I, J | K \rangle \in \mathcal{T}(N)$. If $C \backslash IJK = \varnothing$, $K \backslash ABC = \varnothing$, $A \cap I \neq \varnothing$, and $(J \backslash C) \cup (B \cap IJK) \neq \varnothing$, then $u \star v$ is defined as*

$$u \star v = \langle A \cap I, (J \backslash C) \cup (B \cap IJK) | C \cup (A \cap K) \rangle.$$

*Otherwise, u ⋆ v is undefined.*

The ⋆-operator for constructing new triplets has some interesting properties. A semi-graphoid independence relation is closed under the ⋆-operator: since for $u$, $v \in \mathcal{I}$, the ⋆-operator basically applies the contraction axiom to two triplets that are o-dominated by $u$ and $v$, we have $u \star v \in \mathcal{I}$. The ⋆-operator further defines the set of *all* o-dominant triplets for an independence relation.

**Lemma 4.2** *Let $\mathcal{D} \subseteq \mathcal{T}(N)$ be a set of triplets that satisfies the following two properties:*

- $\forall_{u \in \mathcal{D}} : \mathrm{sym}(u) \in \mathcal{D}$;
- $\forall_{u,v \in \mathcal{D}} : \text{if } u \star v \text{ is defined, then } \exists_{w \in \mathcal{D}} : u \star v \prec w.$

*Then, the set $\mathcal{I} = \{ u \in \mathcal{T}(N) \mid \exists_{v \in \mathcal{D}} : u \prec v \}$ is a semi-graphoid independence relation on $N$.*

For a proof of the lemma we refer the reader to [7]. A crucial step in the proof is the observation that the contraction axiom preserves o-dominance: if $\langle X, Y | Z \rangle \prec u$ and $\langle X, W | YZ \rangle \prec v$, then $\langle X, WY | Z \rangle \prec u \star v$. Apart from $u$ and $v$, therefore, $u \star v$ is the only potentially o-dominant statement for $\mathrm{sem}(\{u, v\})$. This property does not hold for s-dominance: $\langle X, Y | Z \rangle \not\prec u$ and $\langle X, W | YZ \rangle \not\prec v$ do not imply $\langle X, WY | Z \rangle \not\prec u \star v$. To obtain a similar property as Lemma 4.2 for s-dominance that can be used in determining the set of all maximally s-dominant triplets, we introduce a new operator. This operator constructs a (potentially) new s-dominant triplet from two given triplets.

**Definition 4.3** *Let $u = \langle A, B | C \rangle$, $v = \langle I, J | K \rangle \in \mathcal{T}(N)$. If $A \cap I \neq \varnothing$ and $(J \backslash C) \cup (B \backslash J) \neq \varnothing$, then $u \diamond v$ is defined as*

$$u \diamond v = \langle A \cap I, (J \backslash C) \cup (B \backslash J) | C \cup (K \backslash B) \rangle.$$

*Otherwise, $u \diamond v$ is undefined.*

The ⋆- and ⋄-operators are depicted schematically in Figure 1; the ⋆-operator is represented on the left, and the ⋄-operator on the right. In these diagrams the first argument $\langle A, B | C \rangle$ of the operator is represented by columns, and the second argument $\langle I, J | K \rangle$ by rows. The set $D$ represents $N \backslash ABC$, and $L = N \backslash IJK$. Each square in the figure represents an intersection of specific sets from the two arguments. For instance, the square at the intersection of column $B$ and row $K$ represents the variables of $N$ that in the first operator argument are allocated to the set $B$ and in the second argument to the set $K$. All the variables of $N$ are allocated to one of the sixteen squares. The results of applying the different operators to the two arguments are depicted by the three different print patterns that are assigned to the squares. The two sets of conditionally independent variables — i.e. the first two sets in the triplets of $u \star v$ and $u \diamond v$ — are denoted by a vertical and a horizontal pattern, respectively, while the conditioning variables are represent by the solid black. Note that, compared to the ⋆-operator, the ⋄-operator does not require $C \backslash IJK = \varnothing$, nor $K \backslash ABC = \varnothing$. If $C \backslash IJK = \varnothing$ and $K \backslash ABC = \varnothing$, however, we have $u \star v \not\prec u \diamond v$.

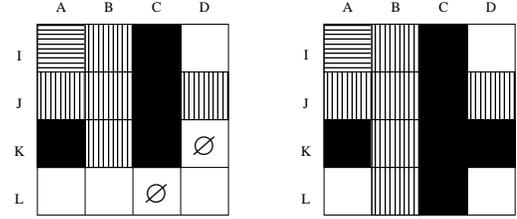

Figure 1: Comparison of the ⋆- and ⋄-operators.

We note that the result of the ⋄-operator is only meaningful when it is applied to triplets from the stable part of an independence relation. When applied to ordinary triplets, the ⋄-operator constructs a new triplet for which we cannot decide if it is in the relation. We now show that the stable part is closed under the operation.

**Lemma 4.4** *Let $\mathcal{I}$ be a stable semi-graphoid independence relation on $N$. If $u$, $v \in \mathcal{I}$, then $u \diamond v \in \mathcal{I}$.*

*Proof.* Let $u = \langle A, B | C \rangle$, $v = \langle I, J | K \rangle \in \mathcal{I}$. Furthermore, let $X = A \cap I$, $Y = B \backslash J$, $Z = C \cup (K \backslash B)$ and $W = J \backslash C$. With these definitions, we find that $\langle X, Y | Z \rangle \not\prec \langle A, B | C \rangle$ and $\langle X, W | YZ \rangle \not\prec \langle I, J | K \rangle$. The triplets $\langle X, Y | Z \rangle$ and $\langle X, W | YZ \rangle$ are therefore in $\mathcal{I}$. The contraction axiom now gives $\langle X, WY | Z \rangle = u \diamond v \in \mathcal{I}$.                    $\square$

With the ⋄-operator we can now define the set of all s-dominant triplets for the stable part of an independence relation.

**Lemma 4.5** *Let $\mathcal{D} \subseteq \mathcal{T}(N)$ be a set of triplets that satisfies the following two properties:*

- $\forall_{u \in \mathcal{D}} : \mathrm{sym}(u) \in \mathcal{D}$;
- $\forall_{u,v \in \mathcal{D}} : \text{if } u \diamond v \text{ is defined, then } \exists_{w \in \mathcal{D}} : u \diamond v \not\prec w.$

*Then, the set $\mathcal{G} = \{ u \in \mathcal{T}(N) \mid \exists_{v \in \mathcal{D}} : u \not\prec v \}$ is a stable semi-graphoid independence relation on $N$.*

*Proof.* From its definition it is immediate that $\mathcal{G}$ is closed under application of the symmetry, decomposition and strong union axioms. It remains to be shown that $\mathcal{G}$ is closed under application of the contraction axiom. Suppose that for some sets of variables $W$, $X$, $Y$, and $Z$, we have $u = \langle X, Y | Z \rangle$, $v = \langle X, W | YZ \rangle \in \mathcal{G}$. By definition there exist a triplet $\langle A, B | C \rangle \in \mathcal{D}$ that s-dominates $u$ and a triplet $\langle I, J | K \rangle \in \mathcal{D}$ that s-dominates $v$. For these s-dominant triplets we have

$$X \subset A, Y \subset B, C \subset Z, \tag{1}$$

$$X \subset I, W \subset J, K \subset YZ. \tag{2}$$

We thus have that $X \subset A \cap I$. Since $X \neq \varnothing$ by definition, we have $A \cap I \neq \varnothing$. Since $W$ and $Z$ are disjoint, moreover, (1) and (2) imply $W \subset J \backslash C$. Since $Y \subset B$, we further have $Y \cap C = \varnothing$. With (1) we find that $Y \subset (B \backslash J) \cup (B \cap J) \subset (B \backslash J) \cup (J \backslash C) \neq \varnothing$. We conclude that $\langle A, B | C \rangle \diamond \langle I, J | K \rangle$ exists. Therefore, there exists a $w \in \mathcal{D}$ that s-dominates this $\diamond$-product. From (1) and (2) we now further find that

$$K \backslash B \subset YZ \backslash B \subset BZ \backslash B = Z \backslash B \subset Z.$$

With $C \subset Z$ we find $C \cup (K \backslash B) \subset Z$. So,

$$\langle X, WY | Z \rangle \prec \langle A, B | C \rangle \diamond \langle I, J | K \rangle \prec w.$$

We conclude that $\langle X, WY | Z \rangle \in \mathcal{G}$ and hence that $\mathcal{G}$ is closed under application of the contraction axiom. $\qquad\square$

# 5 Closure Algorithm

Based upon his concept of o-dominance and Lemma 4.2, Studený [7] defined an algorithm for constructing a compact representation of the semi-graphoid closure of a given set of independence statements. It is based on the idea of repeated application of the $\star$-operator and removal of non-o-dominant triplets. By building upon the $\star$-operator the algorithm has no need to generate the entire closure before selecting the o-dominant triplets that serve to characterise the independence relation. Our algorithm extends on this idea by exploiting the more compact representation for the stable part of the independence relation.

## 5.1 Procedure

The algorithm starts with a set $\mathcal{M}^S$ of stable triplets and a set $\mathcal{M}^U$ of triplets for which stability has not been established. After each iteration, $\mathcal{M}^S$ contains potentially s-dominant triplets of the closure and $\mathcal{M}^U$ contains potentially o-dominant triplets. The triplets between iterations are potentially dominant, in the sense that they have not yet been shown to be dominated by other triplets in the closure.

In each iteration the following steps are performed:

1a: For all $u \in \mathcal{M}^U$, if $\mathrm{sym}(u) \notin \mathcal{M}^U$, then add $\mathrm{sym}(u)$ to $\mathcal{M}^U$.

1b: For all $u \in \mathcal{M}^S$, if $\mathrm{sym}(u) \notin \mathcal{M}^S$, then add $\mathrm{sym}(u)$ to $\mathcal{M}^S$.

2a: For all $u, v \in \mathcal{M}^U$, if $u \star v$ is defined and $u \star v \notin \mathcal{M}^U$, then add $u \star v$ to $\mathcal{M}^U$.

2b: For all $u, v \in \mathcal{M}^S$, if $u \diamond v$ is defined and $u \diamond v \notin \mathcal{M}^S$, then add $u \diamond v$ to $\mathcal{M}^S$.

3a: For all $u = \langle A, B | C \rangle \in \mathcal{M}^U$, $v = \langle I, J | K \rangle \in \mathcal{M}^S$, if $K \backslash ABC = \varnothing$, then add $\langle A, B | C \rangle \star \langle I', J' | K' \rangle$ to $\mathcal{M}^U$ for all triplets $\langle I', J' | K' \rangle \prec \langle I, J | K \rangle$ with $C \backslash I'J'K' = \varnothing$ and $K' \backslash ABC = \varnothing$.

3b: For all $u = \langle A, B | C \rangle \in \mathcal{M}^S$, $v = \langle I, J | K \rangle \in \mathcal{M}^U$, if $C \backslash IJK = \varnothing$, then add $\langle A', B' | C' \rangle \star \langle I, J | K \rangle$ to $\mathcal{M}^U$ for all triplets $\langle A', B' | C' \rangle \prec \langle A, B | C \rangle$ with $C' \backslash IJK = \varnothing$ and $K \backslash A'B'C' = \varnothing$.

4: Check $\mathcal{M}^S$ for new implicit s-dominant triplets.

5a: For all $u \in \mathcal{M}^U$, if there exists a $v \in \mathcal{M}^U \cup \mathcal{M}^S$ that o-dominates $u$, then remove $u$ from $\mathcal{M}^U$.

5b: For all $u \in \mathcal{M}^U \cup \mathcal{M}^S$, if there exists a $v \in \mathcal{M}^S$ that s-dominates $u$, then remove $u$ from $\mathcal{M}^U \cup \mathcal{M}^S$.

The procedure is halted when both $\mathcal{M}^U$ and $\mathcal{M}^S$ remain constant between two iterations.

## 5.2 Elaboration on Step 3

In the closure algorithm of Studený the $\star$-operator is applied to any pair of potentially o-dominant triplets $u$ and $v \in \mathcal{M}$, where $\mathcal{M}$ between iterations is known to include all o-dominant triplets identified so far. In our algorithm, not all such o-dominant triplets are explicitly represented in $\mathcal{M}^U \cup \mathcal{M}^S$: an o-dominant triplet that is s-dominated by a triplet in $\mathcal{M}^S$, is effectively removed from $\mathcal{M}^U$. Step 3 of our algorithm now serves to construct new potentially o-dominant triplets by applying the $\star$-operator to one o-dominant triplet from $\mathcal{M}^U$ that is *not* s-dominated and another o-dominant triplet that *is* s-dominated by a triplet in $\mathcal{M}^S$. Note that the s-dominated triplet is not included in $\mathcal{M}^U$ and therefore needs to be explicitly reconstructed. We further note that it is not necessary to apply the $\star$-operator to any pair of s-dominated triplets, since the result will always be s-dominated by the result of applying the $\diamond$-operator on their s-dominating triplets; this situation is already fully covered by Step 2b of our algorithm.

*Step 3a*

We consider Step 3a, with $u = \langle A, B | C \rangle \in \mathcal{M}^U$ and $v = \langle I, J | K \rangle \in \mathcal{M}^S$. From the previous observations, we have that all o-dominant triplets $v' = \langle I', J' | K' \rangle \prec \langle I, J | K \rangle$ must be reconstructed. For these triplets $u \star v'$ must be calculated and, if defined, the result must be added to $\mathcal{M}^U$. We show that this can be done efficiently.

From the definitions of s-dominance and the $\star$-operator, we have that all triplets $\langle I', J' | K' \rangle$ satisfying

$$I' \subset I, J' \subset J, K' \supset K, \text{ and} \tag{3}$$

$$C \backslash I'J'K' = \varnothing = K' \backslash ABC. \tag{4}$$

must be constructed.

If $C \backslash IJK \neq \varnothing$, then $K'$ has to be a superset of $K_0 = K \cup (C \backslash IJK)$. Note that $K_0$ is the smallest (with respect to set inclusion ordering) superset of $K$ for which $\langle I, J | K_0 \rangle \prec \langle I, J | K \rangle$, $C \backslash IJK_0 = \varnothing$, and $K_0 \backslash ABC = \varnothing$. If $C \backslash IJK = \varnothing$, then $K_0 = K$. For ease of notation,

we now define

$$\langle X, Y | Z \rangle := \langle A, B | C \rangle \star \langle I, J | K_0 \rangle \qquad (5)$$

We address three conditions for $I'$, $J'$ and $K'$ in (3) separately:

• Consider $I' \subset I$. Since $C \backslash I' J K_0$ must be empty, reduction of $I$ to $I'$ is achieved by removing variables from $I \backslash C$. Referring again to Figure 1, we have that removing a variable $r \in I \cap A$ from $I$ amounts to moving $r$ from $I \cap A$ to $L \cap A$. So,

$$\langle A, B | C \rangle \star \langle I \backslash \{r\}, J | K_0 \rangle = \langle X \backslash \{r\}, Y | Z \rangle,$$

which gives an extra triplet to add to $\mathcal{M}^U$. Removing a variable $r \in I \cap B$ from $I$ also results in an extra triplet for $\mathcal{M}^U$:

$$\langle A, B | C \rangle \star \langle I \backslash \{r\}, J | K_0 \rangle = \langle X, Y \backslash \{r\} | Z \rangle.$$

Finally, removing a variable $r \in I \cap D$ does not result in an extra triplet for $\mathcal{M}^U$:

$$\langle A, B | C \rangle \star \langle I \backslash \{r\}, J | K_0 \rangle = \langle X, Y | Z \rangle.$$

• Consider $J' \subset J$. Analogous to the situation above we have that $C \backslash I J' K_0$ must be empty. Reducing $J$ to $J'$ therefore is achieved by removing variables from $J \backslash C$. Removing a variable $r \in J \backslash C$ again gives an extra triplet for $\mathcal{M}^U$:

$$\langle A, B | C \rangle \star \langle I, J \backslash \{r\} | K_0 \rangle = \langle X, Y \backslash \{r\} | Z \rangle.$$

• Finally, we consider $K' \supset K_0$. $K'$ has to satisfy both conditions $C \backslash I J K' = \varnothing$ and $K' \backslash A B C = \varnothing$. This implies that $K$ can be extended from $K_0$ to $K'$ only if a variable $r$ is added from $L \backslash C$. Adding a variable from $L \cap A$ is achieved by moving it from $L \cap A$ to $K \cap A$. This gives an extra triplet to add to $\mathcal{M}^U$:

$$\langle A, B | C \rangle \star \langle I, J | K \cup \{r\} \rangle = \langle X, Y | Z \cup \{r\} \rangle.$$

If we add a variable from $r \in L \cap B$, then this gives an extra triplet for $\mathcal{M}^U$:

$$\langle A, B | C \rangle \star \langle I, J | K \cup \{r\} \rangle = \langle X, Y \cup \{r\} | Z \rangle.$$

Extending $K$ to $K'$ by adding a variable $r \in L \cap D$ is not permitted, since then $K' \backslash A B C$ is no longer empty. It is also possible to extend $K_0$ by moving a variable from $I$ or from $J$ to $K$. The resulting triplet would not be o-dominant, however, since it would be o-dominated by $\langle I, J | K_0 \rangle$. This means that it is not necessary to consider these triplets.

The three cases above for $I'$, $J'$, and $K'$ allow an exact enumeration of *all* o-dominant triplets $v'$ that are s-dominated by $\langle I, J | K_0 \rangle$, and thus also of all new potentially o-dominant triplets $u \star v'$.

*Step 3b*
Under the conditions of Step 3b, all o-dominant triplets that are s-dominated by $\langle A, B | C \rangle$ can be reconstructed in the same way as in Step 3a. First take $C_0 = C \cup (K \cap D)$, to get $K \backslash A B C_0 = \varnothing$. If $\langle X, Y | Z \rangle$ is now defined as $\langle A, B | C_0 \rangle \star \langle I, J | K \rangle$, then the o-dominant triplets are constructed by:

• Reducing $A$ to $A'$: moving $r \in A \cap I$ to $D \cap I$ gives

$$\langle A \backslash \{r\}, B | C_0 \rangle * \langle I, J | K \rangle = \langle X \backslash \{r\}, Y | Z \rangle.$$

Removing a variable $r \in A \cap J$ or $r \in A \cap L$ does not alter $\langle X, Y | Z \rangle$. Removing $r \in A \cap K$ is not allowed, since it leads to $K \backslash A' B C \neq \varnothing$.

• Reducing $B$ to $B'$: moving $r \in B \cap I$ to $D \cap I$ gives

$$\langle A, B \backslash \{r\} | C_0 \rangle * \langle I, J | K \rangle = \langle X, Y \backslash \{r\} | Z \rangle.$$

Removing $r \in B \cap J$ or $B \cap L$ has no effect for the $\star$-operator. Removing $r \in B \cap K$ is not allowed.

• Extending $C_0$: adding $r$ to $C_0$, by moving it from $D \cap I$ to $C \cap I$ gives

$$\langle A, B | C \cup \{r\} \rangle * \langle I, J | K \rangle = \langle X, Y | Z \cup \{r\} \rangle.$$

Adding $r \in D \cap J$ gives

$$\langle A, B | C \cup \{r\} \rangle * \langle I, J | K \rangle = \langle X, Y \backslash \{r\} | Z \cup \{r\} \rangle,$$

but this triplet is o-dominated by $\langle X, Y | Z \rangle$, so it can be omitted. Finally, adding $r \in D \cap K L$ is not permitted. Again in Step 3b it is not necessary to consider extending $C$ by moving variables from $AB$, since this leads to o-dominated triplets.

Also the full set of all o-dominant triplets $\langle A', B' | C' \rangle$ can be enumerated quite efficiently.

Finally, note that in Step 3a we do not add new o-dominant triplets to $\mathcal{M}^U$, for $\langle A, B | C \rangle \in \mathcal{M}^U$, $\langle I, J | K \rangle \in \mathcal{M}^S$, and $C \backslash I J K = \varnothing$, $K \backslash A B C \neq \varnothing$. Since s-dominance allows a search for o-dominant triplets only in $K' \supset K$, it is not possible to find a $K'$, that would make $K' \backslash A B C = \varnothing$. A similar remark applies to Step 3b and the situation when $C \backslash I J K \neq \varnothing$, $K \backslash A B C = \varnothing$.

### 5.3 Elaboration on Step 4

During the loop in the Algorithm of Section 5.1 it can happen that the combination of a number of s-dominant triplets in $\mathcal{M}^S$ can lead to a triplet becoming s-dominant without it being explicitly accounted for in the set $\mathcal{M}^S$. Consider, as an example, the case where $N$ is the union of the disjoint singleton sets $A$, $B$, $C$, $D$, $E$, and $F$ and we have the following three stable independence statements in $\mathcal{M}^S$:

$$\mathcal{S}_\mathcal{I}\langle A, B | CD \rangle, \mathcal{S}_\mathcal{I}\langle A, B | E \rangle, \mathcal{S}_\mathcal{I}\langle A, B | CF \rangle. \qquad (6)$$

The second statement implies $\mathcal{S}_\mathcal{I}\langle A, B | CE \rangle$, and this together with the other two stable independence statement implies $\mathcal{S}_{\mathcal{I}}\langle A, B|C'\rangle$, so the three statements in (6) can be replaced with

$$\mathcal{S}_{\mathcal{I}}\langle A, B|C\rangle, \mathcal{S}_{\mathcal{I}}\langle A, B|E\rangle.$$

This check can be formalised as follows: Assume that $\langle A, B|C\rangle \in \mathcal{M}^S$ and $C \neq \varnothing$. Let $\{d\}$ be a singleton subset of $C$ and $C' := C\backslash\{d\}$. Now $\langle A, B|C'\rangle \in \mathcal{S}_{\mathcal{I}}$ is satisfied iff

$$\forall_{e \in N\backslash ABC'}\exists_{w \in \mathcal{M}^s}: \langle A, B|C' \cup \{e\}\rangle \not\prec w.$$

The complexity of this check is polynomial in $\mathrm{card}(N)$.

### 5.4 Finiteness of the algorithm

The algorithm does indeed construct a representation of the semi-graphoid closure by means of o-dominant and s-dominant triplets. This is proven in the next theorem.

**Theorem 5.1** *Suppose that $\mathcal{M}$ is a given set of independence statements, that can be divided into a given set $\mathcal{M}^S$ of stable independence statements and a set $\mathcal{M}^U$ of ordinary independence statements. The procedure above stops after finitely many iterations and it ends with $\mathcal{M}^S$ a set of s-dominant triplets and $\mathcal{M}^U$ a set of o-dominant triplets that together dominate the semi-graphoid closure of $\mathcal{M}$, that is,*

$$\mathrm{sem}(\mathcal{M}) = \mathcal{I}^U \cup \mathcal{I}^S$$

*where*

$$\begin{aligned}
\mathcal{I}^U &= \{u \in \mathcal{T}(N)|\exists_{v \in \mathcal{M}^U}: u \prec v\}, \\
\mathcal{I}^S &= \{u \in \mathcal{T}(N)|\exists_{v \in \mathcal{M}^s}: u \not\prec v\}.
\end{aligned}$$

*Proof.* The proof follows the lines of [7]. For every iteration $i, i \geq 0$, we let $\mathcal{M}_i^S$ and $\mathcal{M}_i^U$ denote the sets $\mathcal{M}^S$ and $\mathcal{M}^U$ at iteration $i$, respectively. We also define

$$\begin{aligned}
\mathcal{I}_i^S &= \{u \in \mathcal{T}(N) \mid \exists_{v \in \mathcal{M}_i^S}: u \not\prec v\}, \\
\mathcal{I}_i^U &= \{u \in \mathcal{T}(N) \mid \exists_{v \in \mathcal{M}_i^U}: u \prec v\}, \\
\mathcal{I}_i &= \mathcal{I}_i^S \cup \mathcal{I}_i^U.
\end{aligned}$$

It is trivial that $\mathcal{M} \subset \mathcal{I}_i \subset \mathcal{I}_{i+1}$ for $i \geq 0$. Since $\mathcal{T}(N)$ is finite, $\mathcal{I}_i^U \subset \mathcal{I}_{i+1}^U$, and $\mathcal{I}_i^S \subset \mathcal{I}_{i+1}^S$, we have that, for some $i \geq 0, \mathcal{I}_i^U = \mathcal{I}_{i+1}^U$ and $\mathcal{I}_i^S = \mathcal{I}_{i+1}^S$. Since $\mathcal{I}_i^S$ and $\mathcal{I}_i^U$ are uniquely determined by $\mathcal{M}_i^S$ and $\mathcal{M}_i^U$, and vice versa, we conclude that $\mathcal{M}_j^S = \mathcal{M}_{j+1}^S$ and $\mathcal{M}_j^U = \mathcal{M}_{j+1}^U$ for some $j \geq 1$. $\mathcal{I}_j^S \cup \mathcal{M}_j^U$ satisfies the conditions for $\mathcal{D}$ of Lemma 4.2, so

$$\{u \in \mathcal{T}(N) \mid \exists_{v \in \mathcal{I}_j^S \cup \mathcal{M}_j^U}: u \prec v\} = \mathcal{I}_j^S \cup \mathcal{I}_j^U$$

is a semi-graphoid that contains $\mathcal{M}$. By Definition 2.2 we then get $\mathrm{sem}(\mathcal{M}) \subset \mathcal{I}_j^S \cup \mathcal{I}_j^U$. By Lemma 6 from [7] and Lemma 4.4 it can proven by induction on $j$ that $\mathcal{I}_j^S \cup \mathcal{I}_j^U \subset \mathrm{sem}(\mathcal{M})$. $\square$

### 5.5 Complexity of the algorithm

Our closure algorithm is more efficient than the original algorithm presented by Studený [7] in several aspects. First, our representation of independence is more compact. Consider, as an example, the situation where $N$ is the union of the disjoint singleton sets of variables $A$, $B$, $C$, $D$, and $E$. Then, the single stable independence statement $\mathcal{S}_{\mathcal{I}}\langle A, B|\varnothing\rangle$ is equivalent to the following ordinary independence statements

$$\begin{aligned}
&\mathcal{I}\langle A, B|\varnothing\rangle \wedge \mathcal{I}\langle A, B|C\rangle \wedge \mathcal{I}\langle A, B|D\rangle \wedge \ldots \\
&\mathcal{I}\langle A, B|E\rangle \wedge \mathcal{I}\langle A, B|CD\rangle \wedge \mathcal{I}\langle A, B|CE\rangle \wedge \ldots \quad (7) \\
&\mathcal{I}\langle A, B|DE\rangle \wedge \mathcal{I}\langle A, B|CDE\rangle.
\end{aligned}$$

Note that none of these statements is o-dominated. As a result of the more compact representation of independence our algorithm uses less data storage than Studený's algorithm. In the best case the reduction of data storage is $O(2^{\mathrm{card}(N)})$. We return to the complexity of representation in Section 6.

The reduction in data storage for our algorithm leads to a proportional reduction in the number of computation steps to be performed, since for any pair of triplets $u, v \in \mathcal{M}^S$ only $u \diamond v$ needs to be established instead of $u' \star v'$ for all o-dominant $u' \not\prec u$ and $v' \not\prec v$. In the best case this leads to a reduction of $O(2^{\mathrm{card}(N)})$ computation steps.

Also, when a pair of triplets $u \in \mathcal{M}^S$, $v \in \mathcal{M}^U$, are processed (Step 3), the number of computations is reduced, when compared to the original algorithm:

- $u' \star v$ or $v \star u'$ for $u' \not\prec u$ can be derived directly from $u \star v$ and $v \star u$;

- the check if $u' \star v$ or $v \star u'$ are well defined is not necessary, since it is automatic in the "loop" over $u'$;

- the case where $u' \star v$ or $u \star v'$ is o-dominated can be detected beforehand.

In the general case the computational savings can be less than described for the best case. The reduction achieved depends on the presence of stable independence statements in the set $\mathcal{M}^S$ when the procedure is started. Consider an independence statement $u \in \mathcal{M}^U$ that is in fact stable. All statements that can be derived from $u$ by applying the strong union axiom are then included in $\mathcal{M}^U$. The stable statements can in essence be identified as such by checking for the presence of all the possible triplets $u' \not\prec u$ in $\mathcal{M}^U$ and $\mathcal{M}^S$. When performed straightforwardly, this check may require $O(2^{\mathrm{card}(N)})$ time or storage, and therefore it is not included in the current version of our algorithm. The design of a more efficient identification of stable independences in $\mathcal{M}^U$ is the topic of current research. The environment lattices used in ATMSs [1] may well provide a means to this end.

## 6　Complexity of semi-graphoids

The algorithm presented in the previous section results in a representation of a semi-graphoid independence relation by a set of s-dominant triplets and a set of o-dominant triplets. We argued, by means of an example, that the two sets of triplets allow a much more compact representation of an independence relation than a set of o-dominant triplets. We now substantiate this observation by introducing a new definition of complexity for semi-graphoid independence relations.

Studený defines the complexity of a semi-graphoid independence relation $\mathcal{I}$ as

$$\mathrm{com}_{sem}(\mathcal{I}) := \tag{8}$$
$$\min\{\mathrm{card}(\mathcal{D}) \mid \mathcal{D} \subset \mathcal{T}(N), \ \mathrm{sem}(\mathcal{D}) = \mathcal{I}\}.$$

For a stable semi-graphoid independence relation, we now define a new concept of complexity.

**Definition 6.1** *Let $\mathcal{I} \subset \mathcal{T}(N)$ be a* stable *semi-graphoid independence relation, then the* complexity *of $\mathcal{I}$ with respect to the* stab *closure operation is defined as*

$$\mathrm{com}_{stab}(\mathcal{I}) := \\ \min\{\mathrm{card}(\mathcal{D}) \mid \mathcal{D} \subset \mathcal{T}(N), \ \mathrm{stab}(\mathcal{D}) = \mathcal{I}\}.$$

We now present our new definition of complexity for semi-graphoid independence relations.

**Definition 6.2** *Let $\mathcal{I} \subset \mathcal{T}(N)$ be a semi-graphoid independence relation, then the* strong complexity $\mathrm{com}_{strong}$ *of $\mathcal{I}$ is defined as*

$$\mathrm{com}_{strong}(\mathcal{I}) := \\ \min\{\mathrm{card}(\mathcal{C}) + \mathrm{card}(\mathcal{D}) \mid \mathrm{sem}(\mathcal{C}) \cup \mathrm{stab}(\mathcal{D}) = \mathcal{I}\}.$$

The concept of strong complexity exploits the more compact representation of the stable part of a semi-graphoid independence relation. It yields the same complexity value as Studený's concept of complexity, however, for independence relations that do not include a non-trivial stable part.

**Lemma 6.3** *For any semi-graphoid independence relation $\mathcal{I} \subset \mathcal{T}(N)$ we have $\mathrm{com}_{strong}(\mathcal{I}) \leq \mathrm{com}_{sem}(\mathcal{I})$. If $\mathcal{I}$ contains only trivially stable independence statements, then $\mathrm{com}_{strong}(\mathcal{I}) = \mathrm{com}_{sem}(\mathcal{I})$*

*Proof.* Let $\mathcal{I} \subset \mathcal{T}(N)$ be a semi-graphoid. Let $\mathcal{C} \subset \mathcal{T}(N)$ be a set that attains the minimum for $\mathrm{com}_{sem}(\mathcal{I})$. Since $\mathrm{sem}(\mathcal{C}) = \mathcal{I} = \mathrm{sem}(\mathcal{C}) \cup \mathrm{stab}(\varnothing)$, we have $\mathrm{com}_{strong}(\mathcal{I}) \leq \mathrm{card}(\mathcal{C}) + 0 = \mathrm{com}_{sem}(\mathcal{A})$.

If $\mathcal{I}$ does not include a non-trivial stable part, then for any $\mathcal{C} \subset \mathcal{T}(N)$ it is impossible to find a non-empty $\mathcal{D}$, such that $\mathcal{I} = \mathrm{sem}(\mathcal{C}) \cup \mathrm{stab}(\mathcal{D})$. The equality $\mathrm{com}_{strong}(\mathcal{I}) = \mathrm{com}_{sem}(\mathcal{I})$ then follows immediately from the definitions of the two concepts of complexity.　□

## 7　Conclusion

We introduced the concept of stability for semi-graphoid independence relations. Building upon this concept we defined an ordering on independence statements that allows for a representation of the independence relation by means of dominant independence statements. We showed that this representation is more compact than existing representations. We further described an algorithm for determining the set of dominant independence statements. In the near future we plan to develop improvements of our algorithm. In addition we foresee to investigate structural properties of Bayesian networks that derive from the stable part of the independence relation to be represented.

## Acknowledgement

This research was (partly) supported by the Netherlands Organisation for Scientific Research (NWO).